# Overview of the HASOC Subtrack at FIRE 2021: Hate Speech and Offensive Content Identification in English and Indo-Aryan Languages

Thomas Mandl[1], Sandip Modha[2], Gautam Kishore Shahi[3], Hiren Madhu[4], Shrey Satapara[5], Prasenjit Majumder[5], Johannes Schäfer[1], Tharindu Ranasinghe[6], Marcos Zampieri[7], Durgesh Nandini[8] and Amit Kumar Jaiswal[9,10]

[1]*University of Hildesheim, Germany*
[2]*LDRP-ITR, Gandhinagar, India*
[3]*University of Duisburg-Essen, Germany*
[4]*Indian Institute of Science, Bangalore, India*
[5]*DA-IICT, Gandhinagar, India*
[6]*University of Wolverhampton, United Kingdom*
[7]*Rochester Institute of Technology, USA*
[8]*University of Bamberg, Germany*
[9]*University of Bedfordshire, United Kingdom*
[10]*University of Leeds, United Kingdom*

## Abstract
The widespread of offensive content online such as hate speech poses a growing societal problem. AI tools are necessary for supporting the moderation process at online platforms. For the evaluation of these identification tools, continuous experimentation with data sets in different languages are necessary. The HASOC track (Hate Speech and Offensive Content Identification) is dedicated to develop benchmark data for this purpose. This paper presents the HASOC subtrack for English, Hindi, and Marathi. The data set was assembled from Twitter. This subtrack has two sub-tasks. Task A is a binary classification problem (Hate and Not Offensive) offered for all three languages. Task B is a fine-grained classification problem for three classes (HATE) Hate speech, OFFENSIVE and PROFANITY offered for English and Hindi. Overall, 652 runs were submitted by 65 teams. The performance of the best classification algorithms for task A are F1 measures 0.91, 0.78 and 0.83 for Marathi, Hindi and English, respectively. This overview presents the tasks and the data development as well as the detailed results. The systems submitted to the competition applied a variety of technologies. The best performing algorithms were mainly variants of transformer architectures.

## 1. Introduction

There are various types of potentially harmful content in social media such as misinformation and fake news [1], aggression [2], cyber-bullying [3, 4], pejorative language [5], offensive language [6], online extremism [7], to name a few. The automatic identification of problematic content has been receiving significant attention from the AI and NLP communities. In particular, the identification of offensive content, most notably hate speech, has been a growing research area. Within this broad area, various related phenomena have been addressed in isolation such as cyber-bulling, misogyny, aggression, and abuse [8, 9, 10] while some recent work has focused on modeling multiple types of offensive content at once [11, 12].

While research in this area has been gaining momentum [13], there is increasing evidence that social media platforms still struggle to keep up with the demand for technology, particularly for languages other than English [14]. For example, a recent article pointed out that Facebook does not have technology for identifying hate speech in the 22 official languages of India, its biggest market worldwide.[1]

To further contribute to the research in this field, the HASOC 2021 competition contributes with empirically-driven research aiming to find the best methods for the identification of offensive content in social media. In its third edition, HASOC 2021 features re-runs of English and Hindi tasks allowing for better comparison with the results from the editions HASOC 2019 [15] and HAOSC 2020 [16]. Marathi, a Indo-Aryan language similar to Hindi spoken by over 80 million people in India, was added as a new language in HAOSC 2021. A Subtask-2 including conversational hate speech is described in an additional overview paper [17].

## 2. Related Work

This section briefly reviews related research on hate speech identification and data sets created with this goal in mind.

**Current Benchmarks** Recent shared task competitions organised such as TRAC [2], HASOC [18] and OffensEval [19] have presented multiple datasets for hate speech and offensive content identification. While a clear majority of these competitions present English data, several recent shared tasks have created new datasets for various languages such as Greek [20], Danish [21], Mexican Spanish [22], and Turkish [23]. These data sets have influenced the creation of machine learning models to automatically detect offensive content, ranging from SVM models [24] with traditional features to state-of-the-art transformer models [25]. As most of these models typically require training data for each language, it is important to have training data for various languages. Furthermore, one data set per language is not sufficient because the topics of hate speech could change, the potential bias of a data set cannot be easily revealed, and the concept cannot be clearly defined but has a subjective component.

These data sets can be categorised in to two main categories. Data sets such as Offensive Language Detection in Spanish Variants (MeOffendEs@IberLEF 2021) [26] and DEtection of TOXicity in comments In Spanish (DETOXIS) [27] focus on general concepts of offensive content while other data sets are dedicated to more specific topics than general offensive content. A recent data set for Russian which models hate against ethnic groups as a multi-class problem [28] and Guest et al. [29] which has annotated misogyny as a multi-class problem are two recent data sets that focus on specific topics in offensive content identification.

---
[1]https://www.nytimes.com/2021/10/23/technology/facebook-india-misinformation.html

**Annotation for Hate Speech**  The key activity in data set creation is annotation. Human annotators need to decide whether the texts presented to them belong to one of the classes relevant to the task. This process can be organised in different ways. There is no commonly agreed best practice. Some researchers employ a small number of experts [29] or non-experts [15] while others rely on crowd workers [30]. There is a high level of subjectivity associated with the labelling and the class assignment. This can be more serious in cases of systematic bias due to different knowledge levels about issues in society or even about language variants [31]. Also demographic features may lead to bias [32]. Sometimes users of data collections consider some tweets as erroneously labelled. However, it needs to be taken into consideration that the data providers need to follow a consistent protocol and deviations in the opinions about individual tweets are natural. These cases of different options and individual standards form part of any data set because typically, more than one person needs to work on the annotation.

The typical method for measuring annotation quality is that some items are annotated at least twice, and metrics for inter-rater agreement measures the agreement. In cases of low agreement, it is unclear whether the reason behind this is a lack of common understanding between the annotators or the collection contains many dubious cases. One study showed that the agreement is substantially lower than for clear cases [33]. Before starting the annotation, it is not clear how large the portion of dubious cases is. So, even the inter-rater agreement cannot be a guarantee that the annotation is very good.

**Reliability of Data Sets**  Hate speech detection systems are not created only for research but also for real-world applications. It is crucial not just to measure the quality of the classification for one data set but also to analyse how well a system can generalise and be transferred to other data sets. This would be an indicator for a high level of generalisability in realistic scenarios.

Substantial experiments by Fortuna et al. [34] showed that training with one data set and testing with another one can decrease the performance by over 30%. Many potential reasons can be seen as obstacles for the generalisability [35, 36, 37, 38] such as dataset size and annotation quality. However, little is known about their effects. Consequently, the creation of further hate speech data sets is necessary not only for measuring the performance of classifiers but also for the analysis of data sets, the creation processes, and measuring the reliability with new methods.

## 3. HASOC Task Overview and Data Set

The HASOC 2021 dataset is another contribution to the growing body of resources for the analysis of Hate Speech classification. In the following sections, the tasks and the creation process of the data set are described.

## 3.1. Task Definition

This task focuses on Hate speech and Offensive language identification for English, Hindi, and Marathi. Sub-task A is a coarse-grained binary classification in which participating systems are required to classify tweets into two classes, namely: Hate or Offensive (HOF) vs Non-Hate and Non-Offensive (NOT).

- **HOF - Hate and Offensive**: This post contains hate, offensive or profane content.
- **NOT - Non Hate-Offensive**: This post does not contain any Hate Speech, profanity or offensive content. This post contains normal content, statements or anything else. If the utterances are considered to be "normal" and not offending to anyone, they should not be labelled as this could be part of youth language or other language registers.

### 3.1.1. Sub-task B: Identifying Hate, profane and offensive posts (fine-grained)

The second sub-task is a fine-grained classification task offered for English and Hindi. Hate-speech and offensive posts from the sub-task A need to be further classified into the following three categories:

- **HATE - Hate speech**: Posts under this class contain Hate speech content. Ascribing negative attributes or deficiencies to groups of individuals because they are members of a group (e.g. "all poor people are stupid"). These posts includes hateful comments toward groups because of race, political opinion, sexual orientation, gender, social status, health condition or similar.
- **OFFN - Offensive**: Posts under this class contain offensive content. Degrading, dehumanizing or insulting an individual.
- **PRFN - Profane**: These posts contain profane words. Unacceptable language in the absence of insults and abuse. This typically concerns the usage of obscenity, swearwords (Fuck etc.) and cursing (Hell! Damn! etc.).

## 3.2. Data Set Assembly

The sampling of the data set was planned during the time when India was facing the second and extremely hard COVID-19 wave. Therefore, during the sampling process, major topics in social media are highly influenced by COVID-19, and these topics are frequent in the data set [39, 40, 41]. In addition to this, tweets were also sampled about topics related to the brutal post-poll violence in the Indian state West Bengal. Table 1 lists the topics and trending hashtags which were used during the sampling period.

To obtain potentially hateful tweets from the very large corpus of tweets, we have trained a weak classifier based on SVM model with N-gram feature on the HASOC 2019 [42] and 2020 [16] data sets. The purpose of this was to create a weak binary classifier that gives an F1-score around 0.5. We used this classifier to predict labels on the downloaded tweet corpus. We randomly selected tweets classified as HOF (hateful/profane/offensive) by the week classifier. We randomly added 5% of the tweets which were not rated as belonging to the class HOF by the classifier. The main rationale behind this merging process is to ensure that the final data set

| Trending Hashtags | Description of Topics |
| --- | --- |
| #ResignModi | Resignation of PM Modi over COVID-19 crisis in India |
| #ModiKaVaccineJumla | Controversy due to shortage of COVID-19 Vaccine |
| #Murderer_Modi | Death due to shortage of Oxygen attributed to Modi |
| #IndiaCovidCrisis | Brutal second COVID-19 wave in India |
| #TMCTerror | West bengal Post-poll violence. |
| #BengalBurning | West Bengal Post-poll violence. |
| #ChineseWave | Anger on China |
| #chinesevirus | Racist tweets on Chinese |
| #communistvirus | Hashtags trend by right-wing group |
| #covidvaccine | COVID-19 Vaccine |
| #NoVaccinePassports | vaccine passport |
| #chinavirus | Racist tweets on Chinese |
| #wuhanvirus | COVID-19 Origin |
| #islamophobia | Tweets related to hatred against Islam |
| #JusticeForShahabuddin | Death of Controversial Indian politician in India |

**Table 1**
Trending topics from the HASOC data set sample contains a balanced distribution of hateful and non-hateful tweets. We downloaded additional tweets using profane keywords to create an even more balanced data set. Table 2 lists examples for different classes from the data set. The size of data sets for training and testing are shown in Table 3 and Table 4.

The tweets were extracted from Twitter using a targeted sampling approach. All tweets were annotated by at least two annotators. Any conflict between the annotators was resolved by a third annotator. The interrater agreement in subtask 1A is 69% and 72 % for English and Hindi, respectively. For subtask 1B, the agreement for English is 55% and 68% for Hindi.

The data set for Marathi is based on recently released *MOLD* dataset [43]. *MOLD* contains data collected from Twitter. Gaikwad et al. [43] used 22 common curse words in Marathi together with search phrases related to politics, entertainment, and sports along with the hashtag #Marathi. With that, Gaikwad et al. [43] have collected a total 2,547 tweets that were annotated by six volunteer annotators who are native speakers of Marathi. After removing non-Marathi tweets, the final version of *MOLD* contains 2,499 annotated tweets randomly split 75%, 25% into training and testing sets, respectively. Only the sub-task A was available for Marathi.

## 4. Participation and Evaluation

This section details the statistics about the participation in HASOC 2021 by the different teams from all over the world. HASOC 2021 is the third edition of the HASOC at the Forum for Information Retrieval (FIRE) 2021. HASOC started in 2019. This year, HASOC received a record number of participants. A total of 102 teams registered for the participation and 65 teams have submitted 652 runs for all the subtasks. Table 5 summarizes the statistics about the participation.

| Tweet | Task-1 label | Task-2 label |
|---|---|---|
| yeah when she's finally done w you you wanna pop back into her life fuck off | HOF | PRFN |
| #ModiKaVaccineJumla Mr. Modi, where is your "DeshBhakt" BJP workers now??? Do you feel COVID is attacking only the anti-nationals or anti-BJPs ???? Shame a Curse On!!!! | HOF | OFFN |
| @30iPpgStmILw0SI @ChinaDaily #ChineseVirus #WuhanVirus is the #correct name for the #pandemic . #Shameless | NOT | NONE |
| @manoramaonline Shame on people who are still supporting her... including Manorama. keeping MUM #ArrestMamata #BengalBurning #BengalViolence https://t.co/o7lXp6nYZW | HOF | HATE |
| @timotheelvr BITCH GET OUT OF HERE WE ALL KNOW SIALL IS REAL | HOF | PRFN |
| I am booked in to get my first dose of the #Covidvaccine and truth be told I am a bit nervous | First Dog on the Moon https://t.co/u7r8ThfOLW | NOT | NONE |

**Table 2**
Examples of tweet for each class from the data set

| Class | English | Marathi | Hindi |
|---|---|---|---|
| NOT | 1,342 | 1,205 | 3,161 |
| HOF | 2,501 | 669 | 1,433 |
| *PRFN* | 1,196 | - | 213 |
| *HATE* | 683 | - | 566 |
| *OFFN* | 622 | - | 654 |
| Sum | 3,843 | 1,874 | 4,594 |

**Table 3**
Statistical overview of the Training Data

Unlike previously, this year we decided to develop our own submission platform[2] rather than using a third party service. We also provided a leaderboard facility to all participants and the community. The HASOC 2021 leaderboard can be accessed on our Github site[3].

## 5. Results

This section presents the details about the results of the runs by the all participating teams who also submitted a paper describing their system.

Figure 1 presents histograms of the performances of all the teams. Each bin in the histogram depicts a range of 0.01 Macro F1 score. It provides an overview over the distribution of the results.

---
[2]https://hasocfire.github.io/submission/index.html
[3]https://hasocfire.github.io/submission/leaderboard.html

| Class | English | Marathi | Hindi |
|---|---|---|---|
| NOT | 798 | 418 | 1,027 |
| HOF | 483 | 207 | 505 |
| *PRFN* | 379 | | 74 |
| *HATE* | 224 | | 215 |
| *OFFN* | 195 | | 216 |
| Sum | 1,281 | 625 | 1,532 |

Table 4
Statistical overview of the Test Data for determining the final results

| # of teams registered | # of teams submitting runs | # of runs | # of papers |
|---|---|---|---|
| 102 | 65 | 652 | 47 |

Table 5
Participation statistics

| Rank | Team Name | Macro F1 | Rank | Team Name | Macro F1 |
|---|---|---|---|---|---|
| 1 | t1 | 0.7825 | 18 | MUM [44] | 0.7423 |
| 2 | Super Mario [45] | 0.7797 | 19 | BIU [46] | 0.7400 |
| 3 | Hasnuhana | 0.7797 | 20 | Data Pirates [47] | 0.7394 |
| 4 | NLP-CIC | 0.7775 | 21 | TeamBD [48] | 0.7393 |
| 5 | NeuralSpace [49] | 0.7748 | 22 | HNLP [50] | 0.7379 |
| 6 | KuiYongyi [51] | 0.7725 | 23 | JCT | 0.7349 |
| 7 | SATLab [52] | 0.7718 | 24 | TeamOulu [53] | 0.7339 |
| 8 | neuro-utmn-thales [54] | 0.7682 | 25 | SSN_NLP_MLRG [55] | 0.7320 |
| 9 | PreCog IIIT Hyderabad [56] | 0.7648 | 26 | AI-NLP-ML@IITP | 0.7308 |
| 10 | hate-busters | 0.7641 | 27 | Chandigarh_Concordia | 0.7274 |
| 11 | Sakshi HASOC [57] | 0.7612 | 28 | SSNCSE_NLP [58] | 0.7264 |
| 12 | UINSUSKA [59] | 0.7555 | 29 | S_Cube | 0.7195 |
| 13 | IRLab@IITBHU [60] | 0.7547 | 30 | HUNLP [61] | 0.7194 |
| 14 | SOA_NLP [62] | 0.7542 | 31 | TNLP [63] | 0.7181 |
| 15 | algo_unlock [64] | 0.7536 | 32 | IIT_Patna [65] | 0.6848 |
| 16 | UMUTeam [66] | 0.7520 | 33 | JU_PAD [67] | 0.6762 |
| 17 | CAROLL_Passau [68] | 0.7504 | 34 | DLRG [69] | 0.6628 |

Table 6
Results of Task 1A Hindi

## 5.1. Hindi

The best submission for Task A was achieved with a fine-tuned Multilingual-BERT with a classifier layer added at the final phase. The team trained on the HASOC Hindi data set for 20 epochs. With this fine-tuned Multilingual-BERT, the team [45] was able to achieve Macro F1 score of 0.7797.

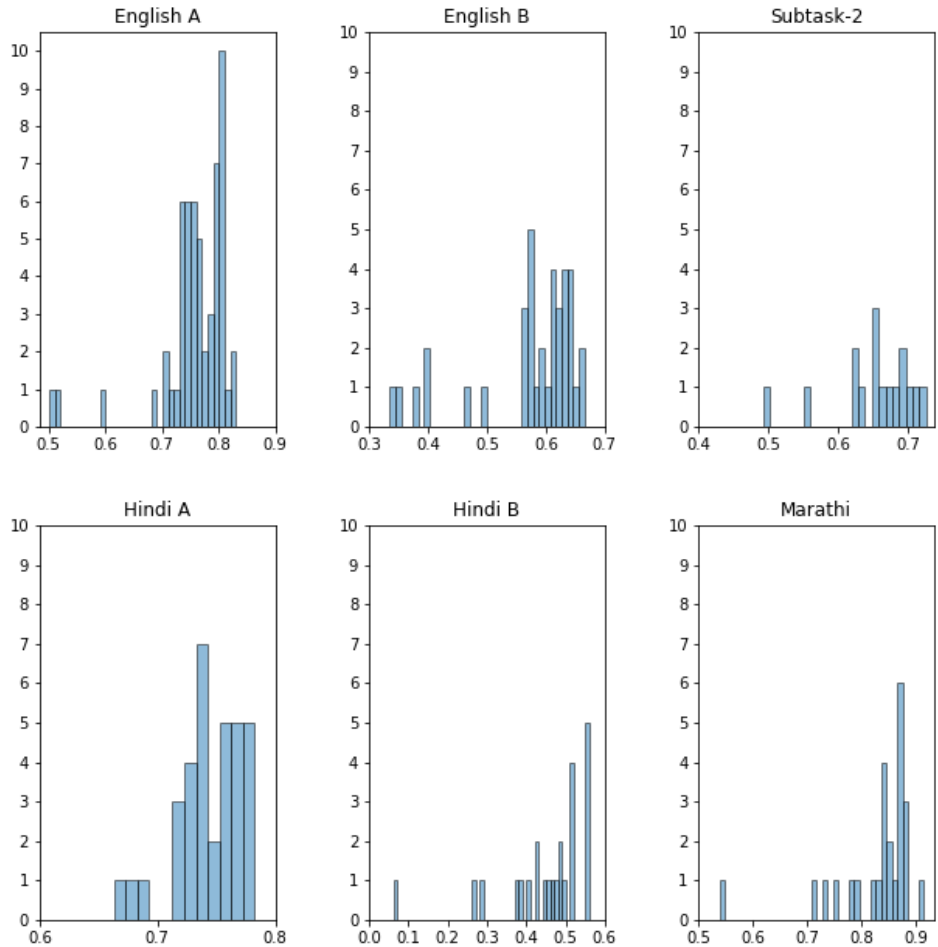

**Figure 1:** Histograms of performance distribution

However, the second team was just 0.0049 points behind this best submission. Apart from fine-tuning a XLM-R transformer, the authors computed vector representations for emojis using the system Emoji2Vec and sentence embeddings for hashtags. These three resulting representations were concatenated before classification. The team was able to achieve the best results for Task B with the same approach [49]. This shows that simply ignoring emojis and hashtags in social media analysis might not always be the adequate approach.

The second team in task B performed just 0.0017 points lower than this best team. This team fine-tuned a Multilingual-BERT transformer with a softmax loss function unlike the two teams previously mentioned which both applied a binary cross Entropy loss.

| Rank | Team Name | Macro F1 | Rank | Team Name | Macro F1 |
| --- | --- | --- | --- | --- | --- |
| 1 | NeuralSpace [49] | 0.5603 | 13 | algo_unlock [64] | 0.4794 |
| 2 | SATLab [52] | 0.5586 | 14 | DLRG [69] | 0.4658 |
| 3 | hate-busters | 0.5582 | 15 | S_Cube | 0.4513 |
| 4 | NLP-CIC | 0.5530 | 16 | HNLP [50] | 0.4431 |
| 5 | KuiYongyi [51] | 0.5509 | 17 | t1 | 0.4290 |
| 6 | UMUTeam [66] | 0.5167 | 18 | UINSUSKA [59] | 0.4257 |
| 7 | IRLab@IITBHU [60] | 0.5127 | 19 | AI-NLP-ML@IITP | 0.4077 |
| 8 | PreCog IIIT Hyderabad [56] | 0.5111 | 20 | Chandigarh_Concordia | 0.3906 |
| 9 | SSN_NLP_MLRG [55] | 0.5110 | 21 | IIT_Patna [65] | 0.3782 |
| 10 | MUM [56] | 0.4952 | 22 | Super Mario [45] | 0.2890 |
| 11 | Data Pirates | 0.4828 | 23 | SOA_NLP [62] | 0.2702 |
| 12 | Hasnuhana | 0.4825 | 24 | Ignite [70] | 0.0621 |

**Table 7**
Results of Task 1B Hindi

Tables 6 and 7 clearly indicate that the top six for Task A and the top five teams for Task B have achieved very close Macro F1s with less than 0.001 difference. For Task A, the mean F1 score achieved by all the best submissions is 0.7436. The standard deviation of the submissions is 0.0289. However, for the top 10 submissions, the standard deviation is only 0.0058. Which is approximately only $\frac{1}{5}$th of the standard deviation of all teams. For task B, the mean F1 score achieved by all the best submissions is 0.4493 which shows that the fine-grained classification remains difficult. We need to consider that the interrater agreement is also low for this task. In this case, the standard deviation between systems is 0.1114, while it is 0.0241 for the best 10 submissions. The standard deviation of all teams is approximately 4.5 times higher than the top 10 teams' standard deviation.

## 5.2. English

The best submission for Task A used a GCN based approach in which the team defined tweets and words as nodes. A word node is connected with all the tweet nodes to which it belongs and a word node is connected to other word nodes that fall into the sliding window of that node across all tweets. Furthermore, the authors used TF-IDF weights as node weights. They were able to achieve 0.8215 as Macro F1 score [61]. The second team used a soft-voting ensemble of four different transformer models jointly fine-tuned on the original training set and the HatebaseTwitter data. using this external ressource, the team was able to achieve a F1 score which is only 0.0016 lower than first team. However the same team ranked first in Task B while using the same approach as for Task A and yielded a Macro F1 of 0.6577 [54]. The second team in Task B used BERT, TF-IDF and the similarity score between the two as features and concatenated them to feed this text representation into a classifier. They achieved a Macro F1 score of 0.6482.

For Task A, the mean F1 score achieved by all the best submissions is 0.7569 while the standard deviation is 0.06255. For the top 10 submissions, the standard deviation is 0.01049 which is approximately $\frac{1}{6}^{th}$ of the standard deviation of all teams. For Task B, the mean F1 score achieved by the best submissions is 0.5707 and while the standard deviation is 0.0888. For the best 10 submissions, the standard deviation is 0.0114. The standard deviation of all teams is approximately 8 times the standard deviation of the top 10 teams.

| Rank | Team Name | Macro F1 | Rank | Team Name | Macro F1 |
|------|-----------|----------|------|-----------|----------|
| 1 | NLP-CIC | 0.8305 | 29 | Alehegn Adane | 0.7623 |
| 2 | HUNLP [61] | 0.8215 | 30 | PC1 | 0.7618 |
| 3 | neuro-utmn-thales [54] | 0.8199 | 31 | TeamBD [48] | 0.7602 |
| 4 | HNLP [50] | 0.8089 | 32 | IIT_Patna [65] | 0.7578 |
| 5 | Chandigarh_Concordia | 0.8040 | 33 | TIB-VA [71] | 0.7565 |
| 6 | KuiYongyi [51] | 0.8030 | 34 | S_Cube | 0.7563 |
| 7 | t1 | 0.8026 | 35 | SOA_NLP [62] | 0.7551 |
| 8 | UINSUSKA [59] | 0.8024 | 36 | SSNCSE_NLP [58] | 0.7541 |
| 9 | TUW-Inf [72] | 0.8018 | 37 | JZ2021 [73] | 0.7497 |
| 10 | UMUTeam [66] | 0.8013 | 38 | Binary Beings [74] | 0.7491 |
| 11 | HASOC21rub [75] | 0.8013 | 39 | E8@IJS | 0.7484 |
| 12 | Super Mario [45] | 0.8006 | 40 | JU_CSE_Team | 0.7468 |
| 13 | Hasnuhana | 0.8006 | 41 | TCS Res. Lab Gurgaon [76] | 0.7448 |
| 14 | NeuralSpace [49] | 0.7996 | 42 | AI-NLP-ML@IITP | 0.7413 |
| 15 | Sakshi HASOC [57] | 0.7993 | 43 | MUM [44] | 0.7389 |
| 16 | IRLab@IITBHU [60] | 0.7976 | 44 | BIU [46] | 0.7388 |
| 17 | PreCog IIIT Hyderabad [56] | 0.7959 | 45 | QQQ [77] | 0.7374 |
| 18 | IMS-SINAI [78] | 0.7947 | 46 | Oswald | 0.7339 |
| 19 | SSN_NLP_MLRG [55] | 0.7919 | 47 | JCT | 0.7327 |
| 20 | giniUs | 0.7909 | 48 | TNLP [63] | 0.7314 |
| 21 | biCourage [79] | 0.7900 | 49 | DLRG [69] | 0.7255 |
| 22 | hate-busters | 0.7894 | 50 | TU Berlin [80] | 0.7203 |
| 23 | SATLab [52] | 0.7823 | 51 | UBCS [81] | 0.7070 |
| 24 | TAD | 0.7776 | 52 | PUCV | 0.7037 |
| 25 | Beware Haters [82] | 0.7722 | 53 | JU_PAD [67] | 0.6813 |
| 26 | TeamOulu [53] | 0.7700 | 54 | NLP_JU | 0.5999 |
| 27 | Vishesh Gupta [83] | 0.7680 | 55 | Team P&P | 0.5133 |
| 28 | AUST_AI | 0.7644 | 56 | ML-LTU | 0.5012 |

Table 8
Results of Task 1A English

### 5.3. Marathi

The best submission for this task use a fine tuned XLM-R Large model with a simple softmax layer to predict the probabilities of class labels. They performed transfer learning from English data released for OffensEval 2019 [19] and Hindi data released for HASOC 2019 [18] and show that performing transfer learning from Hindi is better than performing transfer learning from English. They achieved an F1 score of 0.9144 [84]. Their approach shows the importance of performing transfer learning from a closely related language.

| Rank | Team Name | Macro F1 | Rank | Team Name | Macro F1 |
|---|---|---|---|---|---|
| 1 | NLP-CIC | 0.6657 | 20 | biCourage [79] | 0.5966 |
| 2 | neuro-utmn-thales [54] | 0.6577 | 21 | PreCog IIIT Hyderabad [56] | 0.5927 |
| 3 | HASOC21rub [75] | 0.6482 | 22 | Vishesh Gupta [83] | 0.5871 |
| 4 | Super Mario [45] | 0.6447 | 23 | MUM [56] | 0.5771 |
| 5 | UINSUSKA [59] | 0.6417 | 24 | Binary Beings [74] | 0.5765 |
| 6 | HNLP [50] | 0.6396 | 25 | S_Cube | 0.5739 |
| 7 | Hasnuhana | 0.6392 | 26 | AI-NLP-ML@IITP | 0.5732 |
| 8 | Beware Haters [82] | 0.6311 | 27 | DLRG [69] | 0.5713 |
| 9 | HUNLP [61] | 0.6296 | 28 | giniUs | 0.5666 |
| 10 | UMUTeam | 0.6289 | 29 | IIT_Patna [65] | 0.5652 |
| 11 | NeuralSpace [49] | 0.6268 | 30 | TCS Res. Lab Gurgaon [76] | 0.5638 |
| 12 | SSN_NLP_MLRG | 0.6242 | 31 | TU Berlin [80] | 0.4969 |
| 13 | TUW-Inf [72] | 0.6207 | 32 | Chandigarh_Concordia | 0.4630 |
| 14 | PC1 | 0.6174 | 33 | t1 | 0.4003 |
| 15 | KuiYongyi [51] | 0.6116 | 34 | SOA_NLP [62] | 0.3995 |
| 16 | SATLab [52] | 0.6114 | 35 | QQQ [77] | 0.3770 |
| 17 | hate-busters | 0.6096 | 36 | Team P&P | 0.3454 |
| 18 | IRLab@IITBHU [60] | 0.6093 | 37 | Oswald | 0.3346 |
| 19 | E8@IJS | 0.5994 | | | |

Table 9
Results of Task 1B English

| Rank | Team Name | Macro F1 | Rank | Team Name | Macro F1 |
|---|---|---|---|---|---|
| 1 | WLV-RIT [84] | 0.9144 | 14 | UMUTeam [66] | 0.8423 |
| 2 | neuro-utmn-thales [54] | 0.8808 | 15 | MUM [44] | 0.8411 |
| 3 | Hasnuhana | 0.8756 | 16 | hate-busters | 0.8407 |
| 4 | SATLab [52] | 0.8749 | 17 | Super Mario [45] | 0.8395 |
| 5 | PreCog IIIT Hyderabad [56] | 0.8734 | 18 | Sakshi HASOC [57] | 0.8306 |
| 6 | BIU [46] | 0.8697 | 19 | SSN_NLP_MLRG [55] | 0.8223 |
| 7 | t1 | 0.8696 | 20 | HUNLP [61] | 0.7895 |
| 8 | JCT | 0.8693 | 21 | SSNCSE_NLP [55] | 0.7773 |
| 9 | algo_unlock [64] | 0.8657 | 22 | TNLP [63] | 0.7519 |
| 10 | NeuralSpace [49] | 0.8651 | 23 | DLRG [69] | 0.7338 |
| 11 | KuiYongyi [51] | 0.8611 | 24 | Chandigarh_Concordia | 0.7096 |
| 12 | IRLab@IITBHU [60] | 0.8545 | 25 | Mind Benders [85] | 0.5388 |
| 13 | NLP-CIC | 0.8472 | | | |

Table 10
Results of Task 1A Marathi

The team in second place applied a fine tuned LaBSE transformer [86] on the Marathi data set as well as on the Hindi data set and achieved a F1 score of 0.8808. Their experiments show that LaBSE transformer [86] outperforms XLM-R in the monolingual settings, but XLM-R performs better when Hindi and Marathi data are combined [54].

For task A in Marathi, the mean F1 score achieved by all submissions is 0.8255 and while the standard deviation is 0.0774. Again for the top 10 submissions, the standard deviation is much lower and lies at 0.0143.

## 6. Conclusions and Future Work

The third edition of HASOC has shown that transformer-based classification techniques are the state-of-the-art approach for hate speech and offensive content identification online. This corroborates the findings of recent related competitions such as OffensEval 2020 at SemEval [87]. The best results obtained by participants of HASOC 2021 in terms of macro F1-score were 0.83 in English, 0.78 in Hindi, 0.91 in Marathi. From Figure 1, we can argue that the results can be approximated by a negatively skewed distribution.

In a potential future edition of HASOC, we could encourage participants to use some time-series based classification model for the classification of tweets [88, 89]. HASOC 2021 offered a set of tasks for English, Hindi and Marathi. In the upcoming HASOC edition, we intend to investigate a task for summarization of hateful and normal tweets on long-running debatable topics [90] such as the Middle-East crisis, the Kashmir problem and religious intolerance.

## Acknowledgments

We are thankful to Mr. Pavan Pandya and Mr. Harshil Modh for their contribution in developing the HASOC run submission platform and in the annotation process. We are also thankful to Ms. Mohana Dave and Mr. Vraj Shah for help in the data set sampling and annotation process. We thank all reviewers for HASOC 2021 for their work in a short period of time. We also thank Ms. Ramona Böcker for supporting the paper checking process.

## References

[1] P. Nakov, G. D. S. Martino, T. Elsayed, A. Barrón-Cedeño, R. Míguez, S. Shaar, F. Alam, F. Haouari, M. Hasanain, W. Mansour, B. Hamdan, Z. S. Ali, N. Babulkov, A. Nikolov, G. K. Shahi, J. M. Struß, T. Mandl, M. Kutlu, Y. S. Kartal, Overview of the CLEF-2021 CheckThat! lab on detecting check-worthy claims, previously fact-checked claims, and Fake News, in: Experimental IR Meets Multilinguality, Multimodality, and Interaction - 12th International Conference of the CLEF Association, CLEF Virtual Event, September 21-24, volume 12880 of *Lecture Notes in Computer Science*, Springer, 2021, pp. 264–291. URL: https://doi.org/10.1007/978-3-030-85251-1_19.

[2] R. Kumar, A. K. Ojha, S. Malmasi, M. Zampieri, Evaluating aggression identification in social media, in: Proceedings of the Second Workshop on Trolling, Aggression and Cyberbullying, TRAC@LREC 2020, Marseille, France, May, European Language Resources Association (ELRA), 2020, pp. 1–5. URL: https://aclanthology.org/2020.trac-1.1/.

[3] J. Shetty, K. Chaithali, A. M. Shetty, B. Varsha, V. Puthran, Cyber-bullying detection: A comparative analysis of twitter data, in: Advances in Artificial Intelligence and Data Engineering, Springer, 2020, pp. 841–855. doi:DOIhttps://doi.org/10.1007/978-981-15-3514-7_62.

[4] W. N. H. W. Ali, M. Mohd, F. Fauzi, Identification of profane words in cyberbullying incidents within social networks, Journal of Information Science Theory and Practice 9 (2021) 24–34. doi:https://doi.org/10.1633/JISTaP.2021.9.1.2.

[5] L. P. Dinu, I.-B. Iordache, A. S. Uban, M. Zampieri, A computational exploration of pejorative language in social media, in: Findings of the Association for Computational Linguistics: EMNLP 2021, Association for Computational Linguistics, Punta Cana, Dominican Republic, 2021, pp. 3493–3498. URL: https://aclanthology.org/2021.findings-emnlp.296.

[6] T. Ranasinghe, M. Zampieri, Multilingual offensive language identification with cross-lingual embeddings, in: Proceedings of the 2020 Conference on Empirical Methods in Natural Language Processing (EMNLP), Association for Computational Linguistics, Online, 2020, pp. 5838–5844. URL: https://aclanthology.org/2020.emnlp-main.470. doi:10.18653/v1/2020.emnlp-main.470.

[7] S. Aldera, A. Z. Emam, M. Al-Qurishi, M. A. AlRubaian, A. Alothaim, Online extremism detection in textual content: A systematic literature review, IEEE Access 9 (2021) 42384–42396. URL: https://doi.org/10.1109/ACCESS.2021.3064178.

[8] S. Jaki, T. De Smedt, M. Gwóźdź, R. Panchal, A. Rossa, G. De Pauw, Online hatred of women in the incels. me forum: Linguistic analysis and automatic detection, Journal of Language Aggression and Conflict 7 (2019) 240–268. doi:https://doi.org/10.1075/jlac.00026.jak.

[9] P. Fortuna, J. Soler Company, L. Wanner, Toxic, hateful, offensive or abusive? what are we really classifying? an empirical analysis of hate speech datasets, in: Proceedings of the 12th Language Resources and Evaluation Conference, LREC, Marseille, France, May 11-16, European Language Resources Association, 2020, pp. 6786–6794. URL: https://aclanthology.org/2020.lrec-1.838/.

[10] M. Mladenovic, V. Osmjanski, S. Vujicic Stankovic, Cyber-aggression, cyberbullying, and cyber-grooming: A survey and research challenges, ACM Computing Surveys 54 (2021) 1:1–1:42. URL: https://doi.org/10.1145/3424246.

[11] M. Zampieri, S. Malmasi, P. Nakov, S. Rosenthal, N. Farra, R. Kumar, Predicting the type and target of offensive posts in social media, in: Proceedings of the 2019 Conference of the North American Chapter of the Association for Computational Linguistics: Human Language Technologies, Volume 1 (Long and Short Papers), Association for Computational Linguistics, Minneapolis, Minnesota, 2019, pp. 1415–1420. URL: https://aclanthology.org/N19-1144. doi:10.18653/v1/N19-1144.

[12] S. Rosenthal, P. Atanasova, G. Karadzhov, M. Zampieri, P. Nakov, SOLID: A large-scale semi-supervised dataset for offensive language identification, in: Findings of the Association for Computational Linguistics: ACL-IJCNLP 2021, Association for Computational Linguistics, Online, 2021, pp. 915–928. URL: https://aclanthology.org/2021.findings-acl.80. doi:10.18653/v1/2021.findings-acl.80.

[13] S. Jaki, S. Steiger (Eds.), Digitale Hate Speech - Interdisziplinäre Perspektiven auf Erkennung, Beschreibung und Regulation, Springer, Cham, 2022.

[14] S. Modha, P. Majumder, T. Mandl, C. Mandalia, Detecting and visualizing hate speech in social media: A cyber watchdog for surveillance, Expert Systems and Applications 161

(2020) 113725. URL: https://doi.org/10.1016/j.eswa.2020.113725. doi:10.1016/j.eswa.2020.113725.

[15] T. Mandl, S. Modha, P. Majumder, D. Patel, Overview of the HASOC track at FIRE 2019: Hate Speech and Offensive Content Identification in Indo-European Languages), in: Working Notes of the Annual Meeting of the Forum for Information Retrieval Evaluation, FIRE, CEUR-WS, 2019. URL: http://ceur-ws.org/Vol-2517/T3-1.pdf.

[16] T. Mandl, S. Modha, G. K. Shahi, A. K. Jaiswal, D. Nandini, D. Patel, P. Majumder, J. Schäfer, Overview of the HASOC track at FIRE 2020: Hate speech and offensive content identification in Indo-European Languages, in: Working Notes of FIRE 2020 - Forum for Information Retrieval Evaluation, Hyderabad, India, December 16-20, volume 2826, CEUR-WS.org, 2020, pp. 87–111. URL: http://ceur-ws.org/Vol-2826/T2-1.pdf.

[17] S. Satapara, S. Modha, T. Mandl, H. Madhu, P. Majumder, Overview of the HASOC Subtrack at FIRE 2021: Conversational Hate Speech Detection in Code-mixed language , in: Working Notes of FIRE 2021 - Forum for Information Retrieval Evaluation, CEUR, 2021.

[18] S. Modha, T. Mandl, P. Majumder, D. Patel, Overview of the HASOC track at FIRE 2019: Hate speech and offensive content identification in Indo-European Languages, in: Working Notes of FIRE 2019 - Forum for Information Retrieval Evaluation, Kolkata, India, December 12-15, volume 2517, CEUR-WS.org, 2019, pp. 167–190. URL: http://ceur-ws.org/Vol-2517/T3-1.pdf.

[19] M. Zampieri, S. Malmasi, P. Nakov, S. Rosenthal, N. Farra, R. Kumar, SemEval-2019 task 6: Identifying and categorizing offensive language in social media (OffensEval), in: Proceedings of the 13th International Workshop on Semantic Evaluation, Association for Computational Linguistics, Minneapolis, Minnesota, USA, 2019, pp. 75–86. URL: https://aclanthology.org/S19-2010. doi:10.18653/v1/S19-2010.

[20] Z. Pitenis, M. Zampieri, T. Ranasinghe, Offensive language identification in Greek, in: Proceedings of the 12th Language Resources and Evaluation Conference, European Language Resources Association, Marseille, France, 2020, pp. 5113–5119. URL: https://aclanthology.org/2020.lrec-1.629.

[21] G. I. Sigurbergsson, L. Derczynski, Offensive language and hate speech detection for Danish, in: Proceedings of the 12th Language Resources and Evaluation Conference, European Language Resources Association, Marseille, France, 2020, pp. 3498–3508. URL: https://aclanthology.org/2020.lrec-1.430.

[22] M. E. Aragón, M. Á. Á. Carmona, M. Montes-y Gómez, H. J. Escalante, L. V. Pineda, D. Moctezuma, Overview of MEX-A3T at IberLEF 2019: Authorship and aggressiveness analysis in Mexican Spanish Tweets., in: Iberian Languages Evaluation Forum (IberLEF) SEPLN, 2019, pp. 478–494. URL: http://ceur-ws.org/Vol-2421/MEX-A3T_overview.pdf.

[23] Ç. Çöltekin, A corpus of Turkish offensive language on social media, in: Proceedings of the 12th Language Resources and Evaluation Conference, LREC Marseille, France, May 11-16, European Language Resources Association, 2020, pp. 6174–6184. URL: https://aclanthology.org/2020.lrec-1.758/.

[24] V. Indurthi, B. Syed, M. Shrivastava, N. Chakravartula, M. Gupta, V. Varma, FERMI at SemEval-2019 task 5: Using sentence embeddings to identify hate speech against immigrants and women in Twitter, in: Proceedings of the 13th International Workshop on Semantic Evaluation, Association for Computational Linguistics, Minneapolis, Minnesota,

USA, 2019, pp. 70–74. URL: https://aclanthology.org/S19-2009. doi:10.18653/v1/S19-2009.
[25] T. Ranasinghe, M. Zampieri, H. Hettiarachchi, Brums at hasoc 2019: Deep learning models for multilingual hate speech and offensive language identification, in: FIRE (Working Notes), CEUR, 2019.
[26] F. M. Plaza-del Arco, M. Casavantes, H. J. Escalante, M. T. Martín-Valdivia, A. Montejo-Ráez, M. Montes, H. Jarquín-Vásquez, L. Villaseñor-Pineda, et al., Overview of MeOffendEs at IberLEF 2021: Offensive language detection in Spanish variants, Procesamiento del Lenguaje Natural 67 (2021) 183–194. URL: http://journal.sepln.org/sepln/ojs/ojs/index.php/pln/article/view/6388.
[27] J. Gonzalo, M. Montes-y-Gómez, P. Rosso, Iberlef 2021 overview: Natural language processing for iberian languages, in: Proceedings of the Iberian Languages Evaluation Forum (IberLEF 2021) co-located with the Conference of the Spanish Society for Natural Language Processing (SEPLN 2021), XXXVII International Conference of the Spanish Society for Natural Language Processing., Málaga, Spain, September, 2021, volume 2943 of *CEUR Workshop Proceedings*, CEUR-WS.org, 2021, pp. 1–15. URL: http://ceur-ws.org/Vol-2943/Overview_iberLEF_2021.pdf.
[28] E. V. Pronoza, P. Panicheva, O. Koltsova, P. Rosso, Detecting ethnicity-targeted hate speech in Russian social media texts, Information Processing and Management 58 (2021) 102674. URL: https://doi.org/10.1016/j.ipm.2021.102674.
[29] E. Guest, B. Vidgen, A. Mittos, N. Sastry, G. Tyson, H. Z. Margetts, An expert annotated dataset for the detection of online misogyny, in: Proceedings of the 16th Conference of the European Chapter of the Association for Computational Linguistics: Main Volume, EACL 2021, Online, April 19 - 23, 2021, Association for Computational Linguistics, 2021, pp. 1336–1350. URL: https://aclanthology.org/2021.eacl-main.114/.
[30] J. Pavlopoulos, J. Sorensen, L. Laugier, I. Androutsopoulos, SemEval-2021 task 5: Toxic spans detection, in: Proceedings of the 15th International Workshop on Semantic Evaluation (SemEval-2021), Association for Computational Linguistics, Online, 2021. URL: https://aclanthology.org/2021.semeval-1.6.
[31] M. Sap, D. Card, S. Gabriel, Y. Choi, N. A. Smith, The risk of racial bias in hate speech detection, in: Proceedings of the 57th Conference of the Association for Computational Linguistics, ACL, Florence, Italy, July 28- August 2, Volume 1: Long Papers, Association for Computational Linguistics, 2019, pp. 1668–1678. URL: https://doi.org/10.18653/v1/p19-1163.
[32] H. A. Kuwatly, M. Wich, G. Groh, Identifying and measuring annotator bias based on annotators' demographic characteristics, in: Proceedings of the Fourth Workshop on Online Abuse and Harms, WOAH Online, November 20, Association for Computational Linguistics, 2020, pp. 184–190. URL: https://doi.org/10.18653/v1/2020.alw-1.21.
[33] J. Salminen, H. Almerekhi, A. M. Kamel, S. Jung, B. J. Jansen, Online hate ratings vary by extremes: A statistical analysis, in: Proceedings of the 2019 Conference on Human Information Interaction and Retrieval, CHIIR , Glasgow, Scotland, UK, March 10-14, ACM, 2019, pp. 213–217. doi:10.1145/3295750.3298954.
[34] P. Fortuna, J. Soler Company, L. Wanner, How well do hate speech, toxicity, abusive and offensive language classification models generalize across datasets?, Information Processing and Management 58 (2021) 102524. doi:https://doi.org/10.1016/j.ipm.2021.

102524.

[35] W. Yin, A. Zubiaga, Towards generalisable hate speech detection: a review on obstacles and solutions, PeerJ Computer Science 7 (2021) e598. doi:10.7717/peerj-cs.598.

[36] B. Vidgen, L. Derczynski, Directions in abusive language training data, a systematic review: Garbage in, garbage out, PLOS ONE 15 (2021) 1–32. URL: https://doi.org/10.1371/journal.pone.0243300.

[37] T. Ranasinghe, M. Zampieri, An evaluation of multilingual offensive language identification methods for the languages of india, Information 12 (2021). URL: https://www.mdpi.com/2078-2489/12/8/306. doi:10.3390/info12080306.

[38] T. Ranasinghe, M. Zampieri, Multilingual offensive language identification for low-resource languages, ACM Transactions on Asian and Low-Resource Language Information Processing 21 (2021). URL: https://doi.org/10.1145/3457610.

[39] G. K. Shahi, A. Dirkson, T. A. Majchrzak, An exploratory study of covid-19 misinformation on twitter, Online social networks and media 22 (2021) 100104.

[40] G. K. Shahi, Amused: An annotation framework of multi-modal social media data, arXiv preprint arXiv:2010.00502 (2020).

[41] G. K. Shahi, D. Nandini, Fakecovid–a multilingual cross-domain fact check news dataset for covid-19, arXiv preprint arXiv:2006.11343 (2020).

[42] T. Mandl, S. Modha, P. Majumder, D. Patel, M. Dave, C. Mandlia, A. Patel, Overview of the HASOC track at FIRE 2019: Hate speech and offensive content identification in Indo-European languages, in: Proceedings of the 11th Forum for Information Retrieval Evaluation, 2019, pp. 14–17.

[43] S. S. Gaikwad, T. Ranasinghe, M. Zampieri, C. Homan, Cross-lingual offensive language identification for low resource languages: The case of Marathi, in: Proceedings of the International Conference on Recent Advances in Natural Language Processing (RANLP), Held Online, 1-3 September, 2021, pp. 437–443. URL: https://aclanthology.org/2021.ranlp-1.50.

[44] A. Hegde, M. D. Anusha, H. L. Shashirekha, Ensemble Based Machine Learning Models for Hate Speech and Offensive Content Identification, in: Forum for Information Retrieval Evaluation (Working Notes) (FIRE), CEUR-WS.org, 2021.

[45] S. Banerjee, M. Sarkar, N. Agrawal, P. Saha, M. Das, Exploring Transformer Based Models to Identify Hate Speech and Offensive Content in English and Indo-Aryan Languages, in: Forum for Information Retrieval Evaluation (Working Notes) (FIRE), CEUR-WS.org, 2021.

[46] Y. Hacohen-Kerner, M. Uzan, Detecting Offensive Language in English, Hindi, and Marathi using Classical Supervised Machine Learning Methods and Word/Char N-grams, in: Forum for Information Retrieval Evaluation (Working Notes) (FIRE), CEUR-WS.org, 2021.

[47] P. Mankar, A. Gangurde, D. Chaudhari, A. Pawar, Machine Learning Models for Hate Speech and Offensive Language Identification for Indo-Aryan Language: Hindi, in: Forum for Information Retrieval Evaluation (Working Notes) (FIRE), CEUR-WS.org, 2021.

[48] M. S. Jahan, M. Oussalah, J. K. Mim, M. Islam, Offensive Language Identification Using Hindi-English Code-Mixed Tweets, and Code-Mixed Data Augmentation, in: Forum for Information Retrieval Evaluation (Working Notes) (FIRE), CEUR-WS.org, 2021.

[49] M. Bhatia, T. S. Bhotia, A. Agarwal, P. Ramesh, S. Gupta, K. Shridhar, F. Laumann, A. Dash, One to Rule Them All: Towards Joint Indic Language Hate Speech Detection, in: Forum

for Information Retrieval Evaluation (Working Notes) (FIRE), CEUR-WS.org, 2021.

[50] A. Mitra, P. Sankhala, Multilingual Hate Speech and Offensive Content Detection using Modified Cross-entropy Loss, in: Forum for Information Retrieval Evaluation (Working Notes) (FIRE), CEUR-WS.org, 2021.

[51] Y. Kui, Detect Hate and Offensive Content in English and Indo-Aryan Languages based on Transformer, in: Forum for Information Retrieval Evaluation (Working Notes) (FIRE), CEUR-WS.org, 2021.

[52] Y. Bestgen, A simple language-agnostic yet strong baseline system for hate speech and offensive content identification, in: Forum for Information Retrieval Evaluation (Working Notes) (FIRE), CEUR-WS.org, 2021.

[53] M. S. Jahan, D. R. Beddiar, M. Oussalah, N. Arhab, Y. Bounab, Hate and Offensive language detection using BERT for English Subtask A, in: Forum for Information Retrieval Evaluation (Working Notes) (FIRE), CEUR-WS.org, 2021.

[54] A. Glazkova, M. Kadantsev, M. Glazkov, Fine-tuning of Pre-trained Transformers for Hate, Offensive, and Profane Content Detection in English and Marathi, in: Forum for Information Retrieval Evaluation (Working Notes) (FIRE), CEUR-WS.org, 2021.

[55] K. Adaikkan, T. Durairaj, Multilingual Hate speech and Offensive language detection in English, Hindi, and Marathi languages, in: Forum for Information Retrieval Evaluation (Working Notes) (FIRE), CEUR-WS.org, 2021.

[56] A. Kadam, A. Goel, J. Jain, J. S. Kalra, M. Subramanian, M. Reddy, P. Kodali, A. T. H, M. Shrivastava, P. Kumaraguru, Battling Hateful Content in Indic Languages HASOC '21, in: Forum for Information Retrieval Evaluation (Working Notes) (FIRE), CEUR-WS.org, 2021.

[57] S. Kalra, K. N. Inani, Y. Sharma, G. S. Chauhan, Detection of Hate, Offensive and Profane Content from the Post of Twitter using Transformer-Based Models, in: Forum for Information Retrieval Evaluation (Working Notes) (FIRE), CEUR-WS.org, 2021.

[58] A. Anand, J. Golecha, B. B, B. Jayaraman, M. T. T, Machine Learning based hate speech identification for English and Indo-Aryan languages, in: Forum for Information Retrieval Evaluation (Working Notes) (FIRE), CEUR-WS.org, 2021.

[59] S. Agustian, R. Saputra, A. Fadhilah, "Feature Selection" with Pretrained-BERT for Hate Speech and Offensive Content Identification in English and Hindi Languages, in: Forum for Information Retrieval Evaluation (Working Notes) (FIRE), CEUR-WS.org, 2021.

[60] S. Chanda, S. Ujjwal, S. Das, S. Pal, Fine-tuning Pre-Trained Transformer based model for Hate Speech and Offensive Content Identification in English, Indo-Aryan and Code-Mixed (English-Hindi) languages, in: Forum for Information Retrieval Evaluation (Working Notes) (FIRE), CEUR-WS.org, 2021.

[61] N. Bölücü, P. Canbay, Hate Speech and Offensive Content Identification with Graph Convolutional Networks, in: Forum for Information Retrieval Evaluation (Working Notes) (FIRE), CEUR-WS.org, 2021.

[62] A. Kumar, P. K. Roy, S. Saumya, An Ensemble Approach for Hate and Offensive Language Identification in English and Indo-Aryan Languages , in: Forum for Information Retrieval Evaluation (Working Notes) (FIRE), CEUR-WS.org, 2021.

[63] R. Rajalakshmi, S. Srivarshan, F. Mattins, K. E, P. Seshadri, A. K. M, Conversational Hate-Speech detection in Code-Mixed Hindi-English Tweets, in: Forum for Information

Retrieval Evaluation (Working Notes) (FIRE), CEUR-WS.org, 2021.
[64] A. Velankar, H. Patil, A. Gore, S. Salunke, R. Joshi, Hate and Offensive Speech Detection in Hindi and Marathi, in: Forum for Information Retrieval Evaluation (Working Notes) (FIRE), CEUR-WS.org, 2021.
[65] K. Maity, A. Kumar, S. Saha, Attention Based BERT-FastText model for Hate Speech and Offensive Content Identification in English and Hindi Languages, in: Forum for Information Retrieval Evaluation (Working Notes) (FIRE), CEUR-WS.org, 2021.
[66] C. Caparrós-Laiz, J. Antonio, G. Díaz, R. Valencia-Garcia, Detecting Hate Speech on English and Indo-Aryan Languages with BERT and Ensemble learning, in: Forum for Information Retrieval Evaluation (Working Notes) (FIRE), CEUR-WS.org, 2021.
[67] P. Nandi, D. Das, Detection of Hate or Offensive Phrase using Magnified Tf-Idf, in: Forum for Information Retrieval Evaluation (Working Notes) (FIRE), CEUR-WS.org, 2021.
[68] S. Kannan, J. Mitrović, Hatespeech and Offensive Content Detection in Hindi Language using C-BiGRU, in: Forum for Information Retrieval Evaluation (Working Notes) (FIRE), CEUR-WS.org, 2021.
[69] R. Rajalakshmi, F. Mattins, S. S, P. Reddy, A. K. M, Hate Speech and Offensive Content Identification in Hindi and Marathi Language Tweets using Ensemble Techniques, in: Forum for Information Retrieval Evaluation (Working Notes) (FIRE), CEUR-WS.org, 2021.
[70] I. Jadhav, A. Kanade, V. Waghmare, D. Chaudhari, Hate and Offensive Speech Detection in Hindi Twitter Corpus, in: Forum for Information Retrieval Evaluation (Working Notes) (FIRE), CEUR-WS.org, 2021.
[71] S. Hakimov, R. Ewerth, Combining Textual Features for the Detection of Hateful and Offensive Language, in: Forum for Information Retrieval Evaluation (Working Notes) (FIRE), CEUR-WS.org, 2021.
[72] K. Gémes, A. Kovács, M. Reichel, G. Recski, Offensive text detection on English Twitter with deep learning models and rule-based systems, in: Forum for Information Retrieval Evaluation (Working Notes) (FIRE), CEUR-WS.org, 2021.
[73] J. Zeng, L. Xu, ALBERT for Hate Speech and Offensive Content Identification, in: Forum for Information Retrieval Evaluation (Working Notes) (FIRE), CEUR-WS.org, 2021.
[74] S. Saseendran, S. R, S. V, S. Giri, Classification of Hate Speech and Offensive Content using an approach based on DistilBERT, in: Forum for Information Retrieval Evaluation (Working Notes) (FIRE), CEUR-WS.org, 2021.
[75] W. Yu, B. Boenninghoff, D. Kolossa, Hybrid Representation Fusion for Twitter Hate Speech Identification, in: Forum for Information Retrieval Evaluation (Working Notes) (FIRE), CEUR-WS.org, 2021.
[76] S. Sangwan, L. Dey, M. Shakir, Gated Multi-task learning framework for text classification, in: Forum for Information Retrieval Evaluation (Working Notes) (FIRE), CEUR-WS.org, 2021.
[77] Y. Xu, H. Ning, Y. Sun, Hate Speech and Offensive Content Identification Based on Self-Attention, in: Forum for Information Retrieval Evaluation (Working Notes) (FIRE), CEUR-WS.org, 2021.
[78] F. M. P. del Arco, S. Halat, S. Padó, R. Klinger, Multi-Task Learning with Sentiment, Emotion, and Target Detection to Recognize Hate Speech and Offensive Language, in: Forum for Information Retrieval Evaluation (Working Notes) (FIRE), CEUR-WS.org, 2021.

[79] R. Wilkens, D. Ognibene, biCourage: ngram and syntax GCNs for Hate Speech detection, in: Forum for Information Retrieval Evaluation (Working Notes) (FIRE), CEUR-WS.org, 2021.

[80] S. Mohtaj, V. Schmitt, S. Möller, A Feature Extraction based Model for Hate Speech Identification, in: Forum for Information Retrieval Evaluation (Working Notes) (FIRE), CEUR-WS.org, 2021.

[81] N. P. Motlogelwa, E. Thuma, M. Mudongo, T. Leburu-Dingalo, G. Mosweunyane, Leveraging Text Generated from Emojis for Hate Speech and Offensive Content Identification, in: Forum for Information Retrieval Evaluation (Working Notes) (FIRE), CEUR-WS.org, 2021.

[82] D. N, R. Avireddy, A. Ambalavanan, B. R. Selvamani, Hate Speech Detection using LIME guided Ensemble Method and DistilBERT, in: Forum for Information Retrieval Evaluation (Working Notes) (FIRE), CEUR-WS.org, 2021.

[83] V. Gupta, R. Kumar, R. Pamula, Hate Speech and Offensive Content Identification in English Tweets, in: Forum for Information Retrieval Evaluation (Working Notes) (FIRE), CEUR-WS.org, 2021.

[84] M. Nene, K. North, T. Ranasinghe, M. Zampieri, Transformer Models for Offensive Language Identification in Marathi, in: Forum for Information Retrieval Evaluation (Working Notes) (FIRE), CEUR-WS.org, 2021.

[85] D. Gajbhiye, S. Deshpande, P. Ghante, A. Kale, D. Chaudhari, Machine Learning Models for Hate Speech Identification in Marathi Language, in: Forum for Information Retrieval Evaluation (Working Notes) (FIRE), CEUR-WS.org, 2021.

[86] F. Feng, Y. Yang, D. Cer, N. Arivazhagan, W. Wang, Language-agnostic bert sentence embedding, CoRR abs/2007.01852 (2020). URL: https://arxiv.org/abs/2007.01852.

[87] M. Zampieri, P. Nakov, S. Rosenthal, P. Atanasova, G. Karadzhov, H. Mubarak, L. Derczynski, Z. Pitenis, Ç. Çöltekin, Semeval-2020 task 12: Multilingual offensive language identification in social media (OffensEval 2020), in: Proceedings of the Fourteenth Workshop on Semantic Evaluation, 2020, pp. 1425–1447.

[88] G. K. Shahi, I. Bilbao, E. Capecci, D. Nandini, M. Choukri, N. Kasabov, Analysis, classification and marker discovery of gene expression data with evolving spiking neural networks, in: International Conference on Neural Information Processing, Springer, 2018, pp. 517–527.

[89] D. Nandini, E. Capecci, L. Koefoed, I. Laña, G. K. Shahi, N. Kasabov, Modelling and analysis of temporal gene expression data using spiking neural networks, in: International Conference on Neural Information Processing, Springer, 2018, pp. 571–581.

[90] S. Modha, P. Majumder, T. Mandl, R. Singla, Design and analysis of microblog-based summarization system, Social Network Analysis and Mining 11 (2021) 1–16. URL: https://doi.org/10.1007/s13278-021-00830-3.